%% file: root.tex
\documentclass[letterpaper, 10 pt, conference]{ieeeconf}  

\IEEEoverridecommandlockouts                              

\usepackage[acronym,nonumberlist]{glossaries}
\usepackage{glossaries-prefix}
\usepackage{amssymb}
\usepackage{import}
\newacronym[shortplural=GMMs]{GMM}{GMM}{Gaussian mixture model}
\newacronym[shortplural=HMMs]{HMM}{HMM}{hidden Markov model}
\newacronym[shortplural=DNNs]{DNN}{DNN}{deep neural network}
\newacronym[shortplural=SVDs]{SVD}{SVD}{singular value decomposition}
\newacronym[
    prefixfirst={a\ },
    prefix={an\ }
]{MCTS}{MCTS}{Monte Carlo tree search}
\newacronym[prefixfirst={a\ },prefix={an\ }]{MDP}{MDP}{Markov decision process}
\newacronym{CMDP}{CMDP}{constrained Markov decision process}
\newacronym{RL}{RL}{reinforcement learning}
\newacronym[shortplural=DTs]{DT}{DT}{decision tree}
\newacronym{SMT}{SMT}{satisfiability modulo theories}
\newacronym{IL}{IL}{Imitation Learning}
\newacronym[shortplural=CNNs]{CNN}{CNN}{convolutional neural network}
\newacronym[shortplural=DQNs]{DQN}{DQN}{deep Q-network}
\newacronym{AI}{AI}{artificial intelligence}
\newacronym{PPO}{PPO}{proximal policy optimization}
\newacronym{IG}{IG}{integrated gradients}
\newacronym{PID}{PID}{proportional integral derivative}
\newacronym{MPC}{MPC}{model predictive control}
\usepackage[nocompress,nospace]{cite}
 
\usepackage[usenames,dvipsnames]{xcolor}
\usepackage{pifont}
\usepackage{algorithm}
\usepackage{algpseudocode, amssymb}
\usepackage{booktabs}
\usepackage{multicol,ragged2e}
\usepackage{lineno,todonotes}
\usepackage{amsmath}
\usepackage{import}
\usepackage{varioref}
\usepackage{subfig}
\let\labelindent\relax
\usepackage{enumitem}
\usepackage{kantlipsum}
\usepackage{tabularx}

\usepackage{hyperref}
\hypersetup{%
linktocpage=true, 
colorlinks=false,
pdfborder={0 0 0}, 
breaklinks=true, pdfpagemode=UseNone, pageanchor=true, pdfpagemode=UseOutlines,%
plainpages=false, bookmarksnumbered, bookmarksopen=true, bookmarksopenlevel=1,%
hypertexnames=true, pdfhighlight=/O,
pdftitle={How to Learn from Risk: Explicit Risk-Utility Reinforcement Learning for Efficient and Safe Driving Strategies},%
pdfauthor={Lukas M. Schmidt},%
pdfsubject={SafeDQN},%
pdfkeywords={},%
pdfcreator={pdfLaTeX},%
pdfproducer={LaTeX with hyperref}%
}

\definecolor{GoldenRod}{rgb}{0.85, 0.65, 0.13}
\definecolor{treecyan}{rgb}{0.262, 0.8, 0.6}
\providecommand{\red} [1]{#1}

\definecolor{britishracinggreen}{rgb}{0.0, 0.5, 0.15}

\title{\LARGE \bf
How to Learn from Risk: Explicit Risk-Utility Reinforcement Learning for Efficient and Safe Driving Strategies 
}

\author{Lukas M. Schmidt$^{1}$, Sebastian Rietsch$^{1}$, 
Axel Plinge$^{1\dagger}$, Bjoern M. Eskofier$^{2}$, and Christopher Mutschler$^{1}$
\thanks{$^{1}$~Fraunhofer IIS, Fraunhofer Institute for Integrated Circuits IIS, Nuremberg, Germany. \texttt{\{firstname\}.\{lastname\}@iis.fraunhofer.de}}
\thanks{$^{2}$~Friedrich-Alexander-Universität Erlangen-Nürnberg (FAU), Erlangen, Germany. \texttt{bjoern.eskofier@fau.de}}
\thanks{$^\dagger$ Corresponding Author}
}%
\begin{document}

\maketitle
\thispagestyle{empty}
\pagestyle{empty}

\begin{abstract}
Autonomous driving has the potential to revolutionize mobility and is hence an active area of research. In practice, the behavior of autonomous vehicles must be acceptable, i.e., efficient, safe, and interpretable. While vanilla \gls{RL} finds performant behavioral strategies, they are often unsafe and uninterpretable. Safety is introduced through Safe \gls{RL} approaches, but they still mostly remain uninterpretable as the learned behavior is jointly optimized for safety and performance without modeling them separately. Interpretable machine learning is rarely applied to \gls{RL}.

This work proposes SafeDQN, which allows making the behavior of autonomous vehicles safe and interpretable while still being efficient. SafeDQN offers an understandable, semantic trade-off between the expected risk and the utility of actions while being algorithmically transparent.  We show that SafeDQN finds interpretable and safe driving policies for various scenarios and demonstrate how state-of-the-art saliency techniques can help assess risk and utility. 
\end{abstract}

\glsresetall

\section{Introduction}
\label{sec:introduction}

Achieving sustainable mobility in the future requires new modes of transportation. Autonomous vehicles present a huge opportunity by directly relieving drivers and improving, e.g., ride-sharing services. However, despite recent advances in perception and behavioral planning, traffic participation in complex, real scenarios remains an unsolved problem. Moreover, safety constraints and the need to understand and assess driving behaviors for regulatory approval prevent the use of many modern machine learning models.

In \gls{RL}, an agent learns a desired behavior policy~\cite{schmidt2022marl}. Ideal behavior is subject to three criteria. (1) The agent needs to be \textit{efficient} in reaching the goal in time while conserving energy. (2) The behavior must be \textit{safe}, i.e., the agent is not allowed to take actions that place it or others at high risk of collision or other adverse situations. In other words, the agent needs to optimize for performance only while keeping safety constraints satisfied. (3) Policies and algorithms must be \textit{understandable}. This is crucial for human acceptance~\cite{burkart2021survey}, but also mandated by regulatory constraints~\cite{goodman2017european}, and crucial for the certification and validation~\cite{schmidt2021trust}.

On the one hand, \gls{RL} presents an exciting opportunity to learn efficient and performant driving strategies for almost arbitrary driving situations. On the other hand, safety and interpretability do not come along naturally. A large body of work focusing on safe behavior~\cite{achiam2017constrained, bharadhwaj2021conservative,tessler2019reward, stooke2020responsive, menendez2020maneuver} uses non-transparent mechanisms to guarantee safety. Existing algorithms that find safe and interpretable solutions need to rely on simplifications to provide safety guarantees~\cite{alshiekh2018safe, dalal2018safe, schmidt2021trust}. Thus, there is a clear need for algorithms and agents designed explicitly for interpretability and safety.

\begin{figure}[t]
\centering
\small
\def\svgwidth{\columnwidth}
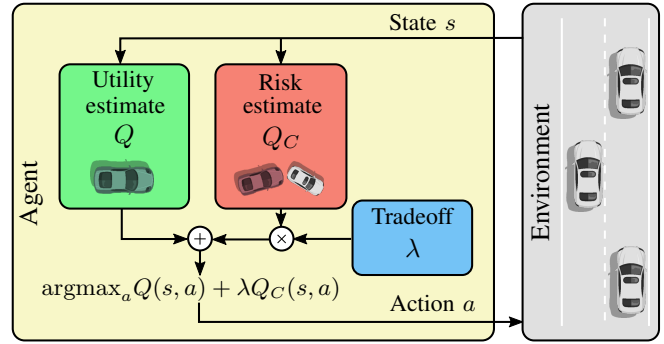
\caption{Overview of SafeDQN.}
\label{fig:summary}
\vspace{-6mm}
\end{figure}

\red{We combine recent advances in Safe \gls{RL} and introduce} SafeDQN, which finds understandable and safe policies. The key idea is a separate estimation of risk and utility in value-based \gls{RL} and their combination into a situation assessment with a learned trade-off parameter, see Fig.~\ref{fig:summary}. In addition to a single $Q$-function, SafeDQN uses an explicit, independent risk estimator $Q_C$, which is trained to estimate the expected discounted constraint costs for each action. Additionally, SafeDQN employs a learned trade-off parameter $\lambda$ similar to RCPO~\cite{tessler2018reward}. Finally, risk and utility estimates are combined for jointly optimal action selection.

The disentangled estimates improve interpretability by allowing understanding and reason about risk and utility separately. We evaluate SafeDQN across a wide range of traffic scenarios and show that it yields efficient strategies while performing much safer than baseline methods. In addition, we demonstrate how our design decisions allow us to gain detailed insights into the operation of our algorithm. This ensures that the internal mechanisms are transparent and understandable.

This paper is structured as follows. Sec.~\ref{sec:basics} introduces \gls{RL} and Safe RL. Sec.~\ref{sec:sota} discusses related work, Sec.~\ref{sec:algorithmic-transparency}  design considerations for safe and interpretable behavior planning, before Sec.~\ref{sec:method} introduces SafeDQN. Sec.~\ref{sec:experiments} evaluates it on different traffic scenarios. Sec.~\ref{sec:interpretability} investigates the interpretability of our policies. Sec.~\ref{sec:conclusion} concludes.

\section{Basics}
\label{sec:basics}
 
Autonomous driving as an instance of sequential decision-making problems can be framed as a \gls{MDP}. An MDP can be defined by a tuple $(S, A, P, r, \gamma)$, where $S$ is a set of states, $A$ is a set of actions, $P = Pr(s' | s, a)$ is the state transition probability (of reaching a state $s'$ from $s$ when taking action $a$), $r(s, a)$ is the reward function, and $0 \leq \gamma \leq 1$ is the discount factor~\cite{sutton1998reinforcement}. We will use $r_t$ for the reward received at time step $t$. In the discounted setting, the goal is to find an optimal policy $\pi^* \in \Pi$ that, given a state $s \in S$, maximizes the expected sum of discounted future reward, i.e., to find
\begin{equation}
    \pi^* = \underset{\pi \in \Pi}{\arg\max} \; \mathbb{E}_{\pi} \; [\sum_{t=0}^{\infty} \gamma^t r(s_t, a_t)].
\end{equation}
The expected, discounted future return for states $s \in S$ under a policy $\pi$ is defined as the value function $V^{\pi}(s)$.

The state-action-value function $Q^{\pi}(s, a)$ gives the expected future return of a state if we take action $a$ first, and then follow the policy $\pi$. To solve the driving task, instead of directly optimizing for continuous or discretized control inputs (e.g., steering, throttle, and brake controls) we let the agent act on a set of high-level, semantic actions, e.g., switch lanes or set and maintain the desired driving speed.

In this work we build upon the \gls{CMDP}~\cite{altman1999constrained} that extends the \gls{MDP} by introducing a cost function $c(s_t, a_t)$ and a constraint $C(s,a)$, which, in expectation and under a given policy $\pi$, must be smaller or equal than a predefined threshold $\vartheta$. We make use of $C$ formulated as a discounted sum, and denote $Q_C(s, a) = \mathbb{E}_{\pi}[C(s,a)]$ as the state-action risk function. More formally, this translates into a constrained optimization problem:
\begin{align}
    \underset{\pi \in \Pi}{\max} \; \; & \mathbb{E}_{\pi} \; [\sum_{t=0}^{\infty} \gamma^t r(s_t, a_t)]\\
    \text{s.t.} \; \; & Q_C^{\pi}(s,a) \leq \vartheta, \; \forall (s,a) \in S \times A\text{.}
\end{align}

A fundamental algorithm in \gls{RL} is Q-Learning. It is an off-policy method that approaches the optimal policy by updating Q-values with the Bellman optimality equation~\cite{sutton1998reinforcement}:
\begin{equation}
    Q^*(s,a) = \mathbb{E}[r_{t+1} + \gamma \max_{a' \in A} Q^*(s_{t+1}, a' | s_t=t, a_t = a)].
\end{equation}

To handle large or non-discrete state spaces, the Q-function can be formulated with a non-linear function approximator such as a neural network (NN)~\cite{mnih2015human}. Hence, the Q-function is defined as $Q(s,a;\theta)$ using weight parameters $\theta$. Such parameters can be learned using the Bellman error of the Q-function (as above) and back-propagation. \red{This idea is key to the Deep Q-Network (DQN)~\cite{mnih2015human} algorithm, which utilizes additional techniques to stabilize learning.}

\section{Related Work}
\label{sec:sota}

A large body of work applies \gls{RL} to the behavioral planning problem in autonomous vehicles. \Gls{RL} has shown to work well across a large variety of scenarios, including highway scenarios~\cite{hoel2018automated, hoel2020reinforcement, ye2020automated,mirchevska2018high,menendez2020maneuver, schmidt2021trust}, roundabouts~\cite{leurent2020safe}, and crash avoidance~\cite{kontes2020high}. They either control steering and velocity directly from sensor readings~\cite{kontes2020high}, or use semantic control actions that are executed using low-level controllers~\cite{mirchevska2018high}. However, such approaches are different from ours as they focus on very few evaluation scenarios, present specific, orthogonal solutions that could be integrated into our work, or add additional constraints on environments.

Another line of research integrates \gls{RL} into planning. Approaches either directly combine RL and \gls{MCTS}~\cite{hoel2020combining} or use a planner to ensure safety during training~\cite{menendez2020maneuver}. This line of work has a large potential but requires accurate forward models to plan with long and useful horizons. In contrast, our model-free approach does not resort to a knowledge of the environment dynamics. 

Safe RL introduces safety constraints into RL optimization problems, most commonly through CMDPs. A large body of work adopts the Lagrangian solution to \glspl{CMDP}. These Lagrangian solutions simplify the problem into a single-objective \gls{RL} problem by adding the constraint costs as a (potentially large) penalty to the rewards. \textit{Reward shaping} often uses a manually selected penalty weight~\cite{schmidt2021trust, ye2020automated} to communicate constraints to the agent's behavior. While this is a relatively simple approach that does not require adaptations to the algorithm, it must be carefully tuned. If the penalty multiplier is too small, reward shaping leads to reward-optimal solutions that violate the constraints.

Advanced Lagrangian methods dynamically adapt the multiplier in response to constraint violations during training. For instance, RCPO~\cite{tessler2019reward} modifies \gls{PPO}~\cite{schulman2017proximal} with the Lagrangian objective and a dynamic multiplier. This dynamic multiplier is increased until the agent no longer violates the constraints to find an optimal trade-off between reward and constraint costs~\cite{tessler2019reward}. Our work adopts this idea in a value-based RL setup.

An additional recent research direction in Lagrangian Safe RL research is the addition of separate estimators for the constraint cost~\cite{geibel2005risk}. CPPO~\cite{stooke2020responsive} uses separate risk and value critics to find optimal updates to a safe policy and performs an alternative update of the trade-off parameter to decrease oscillations. Our algorithm is inspired by these works and adapts their contributions to value-based RL.

Most similar to us, Constrained DQN (CDQN)~\cite{kalweit2020deep} presents an alternate approach to the adaptation. However, CDQN differs from our method in several key aspects:
\begin{enumerate}[leftmargin=*]
    \item In contrast to CDQN which uses a fixed $\beta$-threshold to distinguish between safe and unsafe actions, we exploit the Lagrangian paradigm and learn an optimal trade-off between estimated risk and utility values that satisfies constraints.
    \item We use a different optimization objective that prefers a separate policy when updating $Q$ and $Q_C$. In practice, this is easier (and we found that more complicated constrained objectives~\cite{kalweit2020deep} did not improve performance or safety).
    \item CDQN estimates truncated, undiscounted constraint costs up to a fixed horizon $H$ to improve the interpretability of the estimate. However, in practical scenarios, there is no safe horizon as actions can lead to constraint violations after a long time. Instead, we estimate the discounted expected reward cost. This allows adapting to scenarios with longer risk horizons without harming interpretability.
\end{enumerate}

\section{Algorithmic Transparency}
\label{sec:algorithmic-transparency}

One of the key design considerations for algorithms in autonomous vehicles is algorithmic transparency~\cite{lipton2018mythos}. This means that inputs, internal mechanisms, and outputs of the decision process must be understandable. Algorithmic transparency allows humans to understand the decision-making process. For \gls{RL} in behavioral planning for autonomous vehicles, this results in the following components:

\textbf{Understandable observations and states:} Every element of the input needs to have relevant semantic meaning. Hand-engineered features or affordances~\cite{schmidt2021trust} are inherently more semantic and understandable than raw sensor measurements~\cite{kontes2020high} and are preferred for algorithmic transparency.

\textbf{Interpretable learning framework:} This point specifically addresses the way how to learn from experience. We argue that value-based \gls{RL} methods such as DQNs are inherently easier to interpret than alternative policy-gradient methods such as PPO. DQN iteratively learns to estimate the future rewards that the agent can achieve from a given state-action pair. This value has an exact theoretical definition and can be numerically approximated using sampling. 
In contrast\red{,} policy gradients optimize the policy directly, which works astonishingly well in practice, but is far from intuitive to explain. Further, reducing variance through baselines has an enormous practical effect but introduces additional complexity. While a fully understandable learning framework remains an open research question, we argue that consecutive $Q/Q_C$-updates in SafeDQN are individually explainable.

\textbf{Explicit safety components:} Although the pure Lagrangian formulation and reward shaping have shown to work well, they rely on an implicit trade-off (via the reward definition) between safety and performance. In particular, an external expert cannot verify post hoc that a low $Q$ estimate or action probability corresponds to high estimated risk. SafeDQN makes this trade-off explicit and allows to learn, analyze and verify the risk estimator separately.

\textbf{Semantic actions:} To understand a policy $\pi: S \mapsto A$, we need to fully understand the action space and its semantics, which means that each action must have a clear semantic, understandable meaning. We argue that direct low-level control (where RL directly controls throttle and steering angle), while certainly being effective, cannot be considered fully understandable, because each individual action needs a precise context of other actions to be understood. Instead, research should favor semantic, discrete action spaces. Here, the behavioral planner (an RL agent) has a few discrete, semantic actions available that can be executed using low-level controllers such as \gls{PID} or \gls{MPC}.

\section{Safe Reinforcement Learning with SafeDQN}
\label{sec:method}

SafeDQN takes a different look at the Lagrangian framework and applies it to the DQN framework. We visualize the key ideas in Fig.~\ref{fig:summary} and provide a pseudo code algorithm in Alg.~\ref{alg:safedqn}.

\glspl{CMDP} introduce constraints into the typical setting of an \gls{MDP}. SafeDQN focuses on the setting where constraints are expressed via costs $c_t$ that the agent receives at each time step $t$. This is similar to the reward $r_t$. The constraint is that these costs should not exceed a predefined threshold $\vartheta$.

\textit{Unsafe} \gls{RL} algorithms such as DQN or PPO cannot handle these constraints as they cannot make use of the cost signal during training. As mentioned in Sec.~\ref{sec:sota}, a typical remedy to this is reward shaping, where we fold the costs $c$ into the rewards $r$ using a fixed multiplier $\lambda$ as
\begin{equation}
    \hat{r} = r - \lambda c,
\end{equation}
where $\hat{r}$ is the shaped reward signal fed to the agent. 
Reward shaping essentially penalizes constraint violations and thus makes violating them suboptimal.
However, this fixed penalty coefficient $\lambda$ used in reward shaping is hard to specify, and can often lead to behavior that is too conservative (if $\lambda$ is too high) or too risky (if $\lambda$ is too low). Lagrangian solutions for \glspl{CMDP}, like RCPO, dynamically adapt $\lambda$ during training to find a behavior that exactly satisfies the constraints and achieves optimal reward. This allows the algorithm to increase $\lambda$ while constraints are violated, and to decrease it to find less conservative behaviors once they are satisfied. These solutions are often used with on-policy RL algorithms like PPO, that perform every update on fresh, on-policy transition samples.

This paradigm does not, however, directly translate to off-policy \gls{RL} algorithms, like \gls{DQN}. \Gls{DQN} uses a replay buffer to store environment transitions for the off-policy $Q$-network update. When we naively modify the reward signal and change $\lambda$, these transitions rapidly become stale as they have an incorrect weighting between rewards and costs. This introduces errors into the training data used for updates of the $Q$-network and hinders learning. An alleged solution to this is to store $r$ and $c$ separately inside the replay buffer and to then dynamically apply weighting upon network training, which solves the staleness problem, but still introduces considerable estimation errors to the Q-network whenever $\lambda$ is updated.

\begin{algorithm}[t!]
\caption{SafeDQN.}
\label{alg:safedqn}
\begin{algorithmic}[1]
\Procedure{SafeDQN}{$\text{Env}, \vartheta, \lambda_0, N$}
\State{Initialize estimators $Q$, $Q_C$}
\State{Initialize targets $Q'$, ${Q_C}'$}
\State{Initialize replay buffer $\mathcal{R}$}
\State{Initialize trade-off $\lambda \gets \lambda_0$}
\For{$t \gets 1 \textbf{ to } N$} 
    \State{Store trajectories from Env into $\mathcal{R}$}
    \State{Sample transitions $(s, a, r, c, s')$ from $\mathcal{R}$}
    \State{Calculate $Q$ target $\hat q \gets r + \max Q'(s', a)$}
    \State{Calculate $Q_C$ target $\hat r \gets c + \min {Q_C}'(s', a)$}
    \State{Update $Q$, $Q_C$}
    \State{Update $Q'$, ${Q_C}'$ via polyak averaging}
    \State{Every $\lambda$ update frequency steps:}
    \State{$\lambda \gets \lambda + \alpha \frac{1}{N} \sum_1^N (C_n - \   vartheta)$}
\EndFor
\State{\textbf{return} $Q$, $Q_C$, $\lambda$}
\EndProcedure
\end{algorithmic}
\end{algorithm}

To solve this, SafeDQN trains separate estimators for expected rewards and costs and combines them dynamically using $\lambda$ for rollouts. A state-action value estimator $Q$ corresponds to the typical estimator used by DQN. A risk estimator $Q_C$ is trained to estimate future expected costs incurred when taking an action and then following a risk-optimal policy. We combine these two estimates into a Lagrangian estimate $\hat{Q}$ that is used for the policy as
\begin{equation}
    \hat{Q}(s, a) = Q(s, a) + \lambda Q_C(s, a).
\end{equation}
As in DQN, we use parameterized neural networks $Q$ and $Q_C$ that are trained via replay samples, the Bellmann error, and back-propagation: For a set of transitions, we have two Bellmann optimality equations for utility and risk:
\begin{eqnarray}
    Q^*(s, a) &=& r(s, a) + \gamma \max_{a' \in A} Q^*(s, a') 
\\
    {Q_C}^*(s, a) &=& c(s, a) + \gamma  \min_{a' \in A} Q_C^*(s, a').
\end{eqnarray}
From these, we can calculate the Bellmann error for $Q$ and $Q_C$, and update their weights using minibatch stochastic gradient descent.
For simplicity, we use the same set of transition samples to update $Q$ and $Q_C$.

The trade-off parameter $\lambda$ is updated as
\begin{equation}
    \lambda' = \lambda + \alpha \frac{1}{N} \sum_1^N (C_n - \vartheta),
\end{equation}
where $\alpha$ is a learning rate, $n$ iterates over the last $N$ episodes, $C_n$ is the cumulative cost of constraint violations in episode $n$, and $\vartheta$ is a threshold on the probability of constraint violations that should be upheld. Intuitively, this updates $\lambda$ by adding the probability of constraint violations that exceeds the given threshold, multiplied by the learning rate. We found that updating $\lambda$ with a small frequency (every 2000 steps), as proposed for RCPO~\cite{tessler2019reward}, leads to stable and efficient updates.

\section{Experiments}
\label{sec:experiments}

\begin{figure*}

\centering
\small
\def\svgwidth{\linewidth}
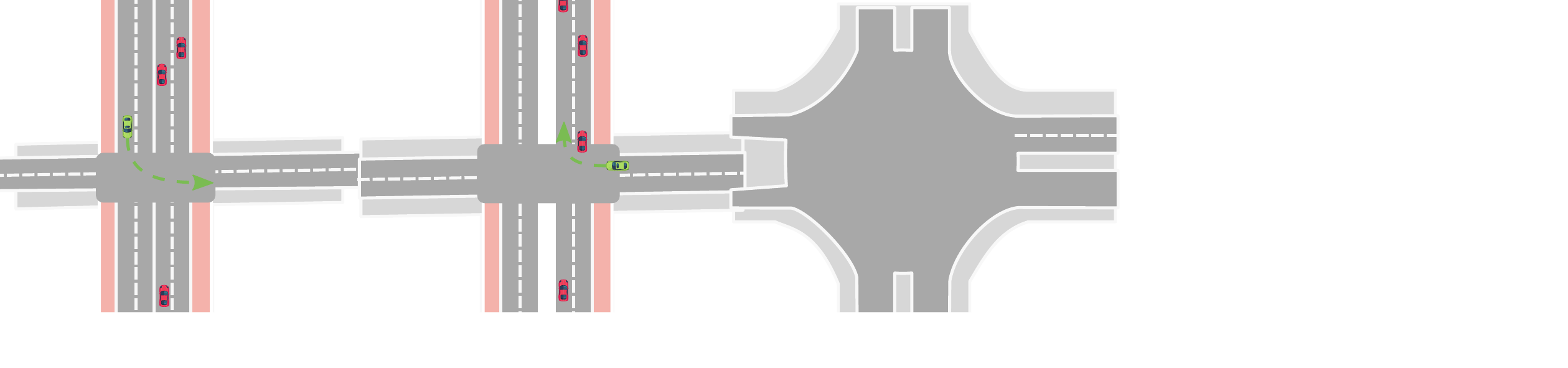
\caption{Visualization of driving environments (ego-vehicle in green).}
\label{fig:scenarios}
\vspace{-4mm}
\end{figure*}

We describe our experimental setup and hyperparameters in Sec.~\ref{sec:hyperparameters}. We evaluate SafeDQN on five different traffic scenarios, see Fig.~\ref{fig:scenarios}. Sec.~\ref{sec:environments} describes these scenarios in more detail. Sec.~\ref{sec:results} discusses the results.

\begin{table}
\caption{Hyperparameters used during training.}
\label{tab:res-hyperparameters}
\centering
\begin{tabular}{lr|lr}
    \toprule
    \multicolumn{4}{c}{All Algorithms} \\
    \multicolumn{2}{l}{$\gamma$} & \multicolumn{2}{r}{0.99} \\
    \multicolumn{2}{l}{ Total training duration} & \multicolumn{2}{r}{2 mio. ts} \\
    \multicolumn{2}{l}{Policy network size} & \multicolumn{2}{r}{2 layers with 256 nodes} \\
    \midrule
    \multicolumn{2}{c}{DQN \& SafeDQN} & \multicolumn{2}{c}{PPO \& RCPO+} \\
    learning rate & 0.001 & learning rate & 0.003 \\
    train frequency & 4 ts & train frequency & 2048\\
    batch size & 32 & batch size & 64 \\
    $\tau$ & 1.0 & epochs & 10 \\
    target update interval & 10\,000 ts & entropy coefficient & 0.0 \\
    gradient ts & 1 & GAE-$\lambda$ & 0.95 \\
    $n$-step estimates & 8 & clip range & 0.2\\
    Exploration \\
    fully random & 50\,000 ts \\ 
    initial $\epsilon$ & 1 \\
    epsilon decay & 200\,000 ts \\
    final $\epsilon$ & 0.05 \\
    \midrule
    \multicolumn{2}{c}{Only SafeDQN} & \multicolumn{2}{c}{Only RCPO+}\\
    Cost threshold $\vartheta$ & 0.001 & Cost threshold & 0.01 \\
    initial trade-off $\lambda_0$ & 100 \\
    Trade-off learning rate $\alpha$ & 1 \\
    $\lambda$ update frequency & 2\,000 ts \\
    \bottomrule
\end{tabular}
\vspace{-4mm}
\end{table}

\subsection{Algorithm and Hyperparameters}
\label{sec:hyperparameters}

Our algorithm implementations are based on an extended version of stable-baselines3~\cite{stable-baselines}, which uses the original implementation of DQNs~\cite{mnih2015human}. We employ multi-step targets~\cite{hessel2018rainbow, sutton1998reinforcement}, i.e., our targets at step $t$ are computed as
\begin{equation}
    \hat Q(s_t, a_t) = \sum_{k = 0}^{n - 1}\gamma^{k}r_{t + k + 1} + \gamma^n \max_{a \in A} Q(s_{t + n}, a).
\end{equation}
We apply these targets for both estimators, replacing $r$ with $c$ and the max-operator with min for the risk estimator. Additionally, we experimented with a dueling-DQN-like decomposition of $Q$ and $Q_C$, but found that it did not improve performance on our tasks. We manually tweaked hyperparameters for DQN and applied the same set for our method and the DQN baseline. Table~\ref{tab:res-hyperparameters} lists all hyperparameters. In particular, we train for 2M time steps in all environments.

\textbf{Baselines:} To evaluate the performance of SafeDQN, we compare it to three baselines:
DQN with reward shaping, PPO with reward shaping, and RCPO+.
RCPO+ is similar to RCPO~\cite{tessler2019reward} but uses two critics (similar to SafeDQN and CPPO~\cite{stooke2020responsive}) that estimate reward and safety separately.

In addition, we evaluated a version of SafeDQN with a different objective function, shown as SafeDQN-AltObjective. In this changed objective we find target actions for both networks with the Lagrangian combination as ${\arg\max}_a Q + \lambda R$. The idea behind this changed objective is to optimize both estimators in a way that includes information from the other. In particular, this changed objective could reduce overestimation in states where the best action (according to $Q$) has a high risk and should not be taken. This idea is similar to Double DQN~\cite{hasselt2016double}, which uses the best action of the target network to reduce overestimation bias. CDQN~\cite{kalweit2020deep} also includes constraints, but in a stricter way.

We ran four repetitions for each algorithm on all environments to account for variance between different seeds. 

\subsection{Environment}
\label{sec:environments}

We evaluate SafeDQN and the baselines on a custom simulation environment based on the open-source traffic simulator SUMO~\cite{SUMO2018,Rietsch2022}. 
In particular, we constructed six driving scenarios of varying complexity as depicted in Fig \ref{fig:scenarios}, where the goal for the agent is to successfully complete each scenario along a predefined route without a collision. We include three typical highway tasks (merge, drive, split) and inner-city scenarios (left/right turns and a roundabout).
In particular, the street networks of scenarios Right-Turn, Left-Turn, and Roundabout are part of the \textit{Town03} map from the CARLA simulator \cite{dosovitskiy2018carla}, which were integrated utilizing SUMO's tooling scripts.

\textbf{Reward:} The driving task uses a reward function
\begin{equation}
    r = r_{\text{dense}} + r_{\text{terminal}}
\end{equation}
that is comprised of a dense driving speed reward
\begin{equation}
    r_{\text{dense}} = v_t / v_{max},
\end{equation}
where $v_t$ is the current driving speed at timestep $t$ and $v_{max}$ the maximum driving speed, and sparse terminal rewards
\begin{equation}
    r_{\text{sparse}} = 
        \begin{cases}
            +100, & \text{if reached goal}\\
            -100, & \text{if crashed or off-route }\\
            0, & \text{otherwise,}
        \end{cases}
\end{equation}
where the environment returns a positive reward of $+100$ if the agent reaches the goal, and a negative reward of -$100$ if the agent leaves the route or collides with another vehicle.\footnote{Note that, due to our action space, the agent cannot leave the streets. Hence, we only need to penalize the agent for "taking a wrong turn".}

\textbf{Constraints:} We treat the collision case as a safety constraint with $c_t=1$ if the agent crashed at time step $t$, and $c_t = 0$ otherwise.
Note that we also include this situation in the reward with a dedicated penalty of -100. This allows even our non-safe baselines to find constraint-satisfying behavior. 

\begin{table*}
\caption{Maximum returns ($R_{max}$) and respective crash rate (C, in \%) averaged across multiple runs for all algorithms and environments, where each policy was evaluated for 100 episodes every 50M timesteps. Data points with crash rates greater than the training run's 10th-percentile are prefiltered beforehand.}
\label{tab:results}
\begin{tabular}{l rr rr rr rr rr rr}
\toprule
 &  \multicolumn{2}{c}{Left-Turn} & \multicolumn{2}{c}{Right-Turn} & \multicolumn{2}{c}{Roundabout} & \multicolumn{2}{c}{Highway-Drive} & \multicolumn{2}{c}{Highway-Merge} & \multicolumn{2}{c}{Highway-Split}\\
\cmidrule{2-3} \cmidrule{4-5} \cmidrule{6-7} \cmidrule{8-9} \cmidrule{10-11} \cmidrule{12-13}
{Algorithm} &         \multicolumn{1}{c}{$R_{max}$} & \multicolumn{1}{c}{C} &       \multicolumn{1}{c}{$R_{max}$} & \multicolumn{1}{c}{C} &        \multicolumn{1}{c}{$R_{max}$} & \multicolumn{1}{c}{C} & \multicolumn{1}{c}{$R_{max}$} & \multicolumn{1}{c}{C} & \multicolumn{1}{c}{$R_{max}$} & \multicolumn{1}{c}{C} & \multicolumn{1}{c}{$R_{max}$} &  \multicolumn{1}{c}{C} \\
\midrule
DQN                  &    78$\pm$78 &    0$\pm$00 &    74$\pm$  4 &   30$\pm$ 26 &    42$\pm$ 31 &    4$\pm$\hphantom{0}6 & 14$\pm$10 &   20$\pm$\hphantom{0}5 &        4$\pm$\hphantom{0}5 &    0$\pm$\hphantom{0}0 &       40$\pm$15 &    0$\pm$\hphantom{0}0 \\
PPO                  &   141$\pm$\hphantom{0}5 &    8$\pm$\hphantom{0}2 &   111$\pm$79 &   10$\pm$\hphantom{0}7 &   118$\pm$11 &   18$\pm$\hphantom{0}1 & 144$\pm$\hphantom{0}7 &   20$\pm$\hphantom{0}3 &       45$\pm$30 &    2$\pm$\hphantom{0}5 &      148$\pm$  4 &    6$\pm$\hphantom{0}1 \\
RCPO+                &   149$\pm$\hphantom{0}3 &    4$\pm$\hphantom{0}2 &    71$\pm$49 &    4$\pm$\hphantom{0}5 &    66$\pm$29 &    6$\pm$\hphantom{0}3 & 153 $\pm$\hphantom{0}4 &   16$\pm$\hphantom{0}2 &       21$\pm$\hphantom{0}2 &    0$\pm$\hphantom{0}0 &       66$\pm$60 &    0$\pm$\hphantom{0}0 \\
SafeDQN              &   150$\pm$\hphantom{0}8 &    0$\pm$\hphantom{0}0 &   156$\pm$47 &    0$\pm$\hphantom{0}0 &    44$\pm$22 &    0$\pm$\hphantom{0}0 & 107$\pm$53 &   18$\pm$\hphantom{0}8 &        5$\pm$17 &    0 $\pm$\hphantom{0}0 &      132$\pm$ 55 &    0$\pm$\hphantom{0}0 \\
SafeDQN-Alt. &           158$\pm$\hphantom{0}1 &    0$\pm$\hphantom{0}0 &   151$\pm$56 &    0$\pm$\hphantom{0}0 &    97$\pm$28 &    0$\pm$\hphantom{0}0 &  96$\pm$31 &   38$\pm$10 &        4$\pm$17 &    0 $\pm$\hphantom{0}0 &       88$\pm$58 &    1$\pm$\hphantom{0}0 \\
\bottomrule
\end{tabular}
\end{table*}
\begin{figure*}
\centering
\small
\def\svgwidth{.99\textwidth}
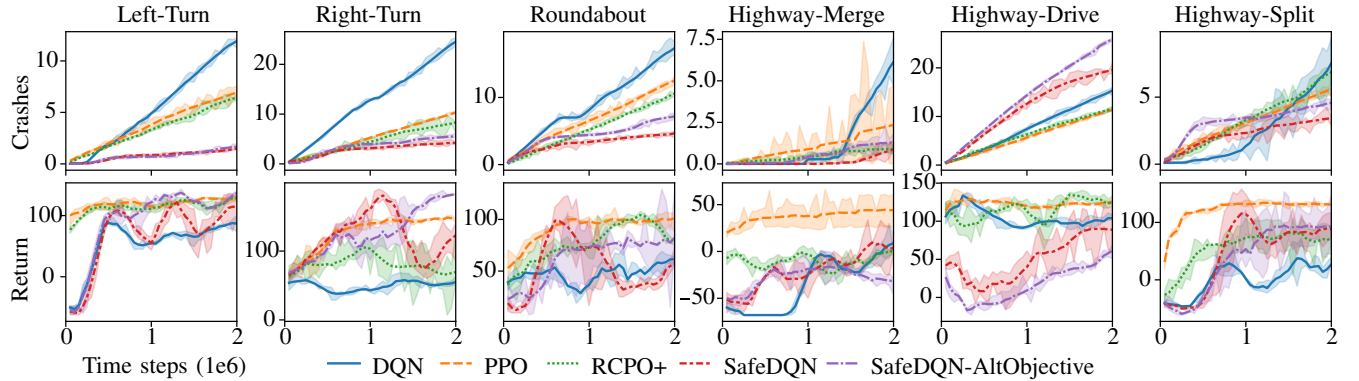
\caption{Training results over simulation time steps. Top row: cumulative evaluation constraint costs; Bottom row: reward.}
\label{fig:training-curves}
\vspace*{-4mm}
\end{figure*} 

\textbf{Action Space:} As mentioned in Sec.~\ref{sec:algorithmic-transparency} a key ingredient for algorithmic transparency in RL are semantic actions. Given the route from start to end goal, we condense the set of possible actions $A$ into actions for switching between lanes, as well as choosing the desired target speed. We discretize the continuous range of possible driving speeds into a finite set via equidistant sampling. In addition, the agent is free to choose a No-Op-action that maintains the current lane and target speed. Formally, the set of actions $A$ is defined as:
\begin{eqnarray}
    A &=&
    \{
         \text{No-Op},
        \text{SwitchLane}_{\text{Left}}, 
        \text{SwitchLane}_\text{Right}, \\&&
        \text{TargetSpeed}_{0},
        \text{TargetSpeed}_{1},
    \ldots
        \text{TargetSpeed}_{N}, \nonumber
    \}
\end{eqnarray}
We realize the semantic actions by sampling a set of lane-centered waypoints from the current lane, which can then be tracked by a low-level controller such as PID or MPC. We resort to a simplified approach, where we adjust the ego vehicle position and rotation w.r.t.\ the next waypoint via linear interpolation under the current driving speed (as we see the integration of a low-level controller as a separate topic). We use $N=5$ different target speeds from $v=0\frac{m}{s}$ to $v = 14\frac{m}{s}$.

\textbf{Observations:} The agent receives observations of the ego vehicle state, non-ego traffic participant information inside a radius around the ego vehicle, the set of current lane waypoints, as well as available lane switching options. The ego state is defined as a vector $[x, y, v, \theta, v_{target}]$, where $x, y$ are the ego coordinates, $v$ is the actual driving speed, $v_{target}$ the desired target speed and $\theta$ the yaw angle. Non-ego vehicle states are defined as vectors $[x', y', v', \varphi, d]$ and are calculated relative to the agent, where $\varphi$ is the rotation angle relative to the ego vehicle and $d$ the distance to the ego vehicle. Waypoint observations are calculated analogously without a value for velocity. Available lane switching options are given as a one-hot encoding. The full observation vector is standardized into the $[-1, 1] \in \mathbb{R}$ range and given as inputs to the respective feed-forward networks.

\textbf{Randomization:} A key ingredient of our environment is different kinds of traffic randomization techniques. These can be summarized as follows:
\begin{enumerate}[leftmargin=*]
    \item Traffic participants are spawned at their predefined scenario entry points based on a probability $p$, which differs between scenarios and entry points and that is sampled uniformly from the range $p \in [0.2, 0.6]$,
    \item the preferred driving speeds of traffic participants (which is the speed limit of the street in the standard setting) is multiplied by a scalar $s \sim \mathcal{N}(\mu=0.7, \sigma=1.0)$ which we clamp into the range $[0.4, 2.0]$,
    \item every $3$ seconds, a random vehicle inside a cone of $60^{\circ}$ in front of the vehicle performs an emergency brake, and
    \item at every timestep, a scalar $s \sim \mathcal{N}(\mu=0, \sigma=1.0)$ multiplied by $5 \frac{m}{s}$ is added to the desired driving speed of a random vehicle inside the cone for $3$ seconds.
\end{enumerate}
We argue that this facilitates better generalization of the agent policy across a wider range of possible traffic scenarios. In particular, we found the addition of random brakes and driving speed changes to yield a noticeable improvement in safety in initial experiments. 

\subsection{Results}
\label{sec:results}
 
In Fig.~\ref{fig:training-curves} we show the cumulative number of constraint violations (top row) and the average return per episode (bottom row) of SafeDQN along with the baselines. As the performance of value-based RL methods is prone to drop after some point during training, we report maximum returns and the respective number of crashes in Table~\ref{tab:results}. Because our principal focus lies on policy safety, we only consider policies with crash rates within the $10$-th percentile of a training run. In general, SafeDQN operates much safer than all baselines on five out of six scenarios (except for Highway-Drive) while achieving competitive performance to PPO at points in training on five scenarios (except for Highway-Merge). The results prove that SafeDQN yields highly competitive behavioral policies while at the same time being algorithmically transparent.

On Highway-Drive, SafeDQN is unable to find a safe policy quickly. Highway-Drive is different from the other scenarios in that a random policy performs competitively. A random policy generally drives at speeds below the traffic's average and random lane changes are easier to compensate for by the traffic's lane-changing model. We hypothesize that this allows PPO to find an adequate policy much quicker by relying on the inherent policy stochasticity, whereas DQN is taking longer to find a deterministic policy. Similar reasoning applies to Highway-Merge, where SafeDQN is unable to find a highly rewarding strategy. This indicates that our action space is not perfectly suited to deterministic policies on highway scenarios.

From our baselines, PPO and RCPO+ outperform classical DQN. This clearly shows that our additions, as well as our additional focus on safety, transform DQN into the highly competitive and safe value-based RL approach.

Our alternate objective (SafeDQN-AltObjective) gives mixed results. It fails to learn an appropriate policy for Highway-Drive and performs slightly less safely. However, we noticed that SafeDQN sometimes oscillates, which is less noticeable with the changed objective. 

\section{Understanding Risk Estimates}
\label{sec:interpretability}
The main benefit of our algorithm is its algorithmic transparency.
The unique structure of our algorithm allows us to understand risk estimates separately from utility estimates, which gives us new tools to assess and validate the generated policies.
To demonstrate this, we evaluate our learned risk estimators in terms of their precision and recall metrics and investigate the importance of specific input values (in particular, of other cars) for the risk assessment.

\subsection{Cost recall and precision}
We first evaluate the quality of our risk estimates. Our risk estimators $Q_C$ are trained to approximate the discounted future sum of costs under the optimal policy $\pi^*$
\begin{equation}
    {Q_C}^*(s, a) = \mathbb{E}_{a \propto \pi^*(s), s \in \mathcal{P}} \sum_{t} \gamma^t c_t.
\end{equation}
During training, however, we do not know $\pi^*$. Also, it is impossible to approximate the expectation via sampling. Instead, we employ two relatively simple metrics inspired by supervised evaluation metrics: cost recall and cost precision.

To define these metrics, we convert the regression task into a classification task by defining a threshold $t$. Actions that have a risk assessment of $Q_C(s, a) > t$ are assumed to be high-risk, whereas actions with lower risk estimates are sorted into a low-risk class. We define the true label for a state-action-pair by the transition cost $c(s,a)$ that occurs directly after taking an action, with $c(s, a) = 1$ belonging to the true-cost class and $c(s, a) = 0$ regarded as no-cost. 

Given these classes, we can now define the cost recall as
\begin{equation}
    \text{Cost Recall} = \frac{n_{c = 1, Q_C > t}}{n_{c = 1, Q_C > t} + n_{c = 1, Q_C \leq t}},
\end{equation}
where $n_X$ is the number of samples in the current replay buffer that satisfy condition $X$. For example, $n_{c=1, Q_C > t}$ defines the number of samples belonging to the true-cost class and having estimated risk values greater than the threshold. The cost recall is a crucial evaluation metric, 
since it measures the fraction of critical high-cost situations that were correctly anticipated by the risk estimator.
A good risk estimator should always adhere to a cost recall close to one. A low cost recall indicates that the estimator often underestimated the risk of actions that lead to immediate violations.

Using the same classes, the cost precision is given by
\begin{equation}
    \text{Cost Precision} = \frac{n_{c = 1, Q_c > t}}{n_{c = 1, Q_c > t} + n_{c = 0, Q_C > t}}.
\end{equation}
The cost precision calculates the number of true positives among the situations that were estimated to have high risk. An accurate estimator should approach a precision of one. A low-cost precision indicates an over-cautious risk estimator that severely overestimates constraint costs on safe actions.

\begin{table}[t!]%
\caption{Cost Recall and Cost Precision results for SafeDQN across all evaluation environments.}%
\label{tab:costrecall}
\centering%
\begin{tabular}{lcc}%
\toprule%
Environment   & Cost Recall & Cost Precision \\%
\midrule%
Left Turn     &  0.96$\pm$0.01 &    0.97$\pm$0.01 \\%
Right Turn    &  0.93$\pm$0.03 &    0.98$\pm$0.01 \\%
Roundabout    &  0.93$\pm$0.01 &    0.98$\pm$0.01 \\%
Highway-Merge &  0.87$\pm$0.07 &    0.88$\pm$0.12 \\%
Highway-Drive &  0.86$\pm$0.04 &    0.21$\pm$0.13 \\%
Highway-Split &  0.81$\pm$0.02 &    0.97$\pm$0.03 \\%
\bottomrule%
\end{tabular}%
\vspace*{-5mm}
\end{table}%

Overall, cost recall and cost precision can give a good overview of potential over- and underestimation errors of the risk estimator. In our experiments, we found that our risk estimators often achieve a very good cost recall (larger than $0.9$ in the city scenarios, and above $0.85$ in the more chaotic highway environments). We found that the cost precision of our estimators improves massively during training and approximated 100\% after around 1M training time steps on all environments except Highway-Drive. Table~\ref{tab:costrecall} summarizes these results. Remarkably, our risk estimators achieve these metrics very reliably, with standard deviations on cost recall and precision lower than $0.05$ in most cases. 

\begin{figure}[b!]%
 \centering%
\small
\def\svgwidth{.88\columnwidth}
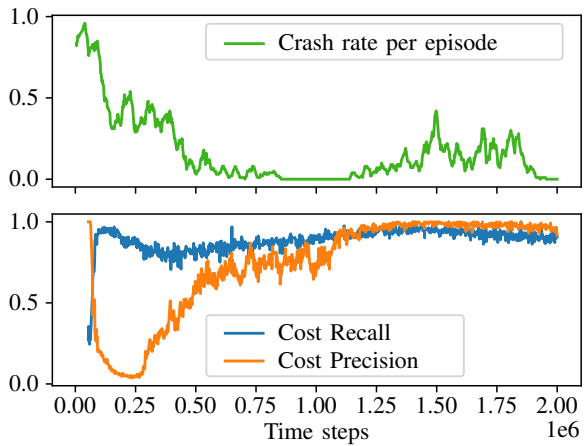
\caption{Cost recall and precision on "Right Turn". }%
\label{fig:recall_precision}%
\end{figure}%
Fig.~\ref{fig:recall_precision} exemplary shows a typical cost recall/precision curve recorded while training SafeDQN on the right turn traffic scenario, along with the costs received during training. Cost recall (bottom, blue curve)
stays at acceptable levels close to 1 throughout the training, with a small drop at 0.5M time steps. Cost precision (bottom, orange curve) quickly climbs to levels close to 1 after around 1M time steps. Note how relatively low-cost precision at the beginning of training correlates with a high number of crashes (top, green curve).

\begin{figure}
    \centering
    \includegraphics[width=\linewidth,trim=0pt 0pt 0pt 0pt,clip]{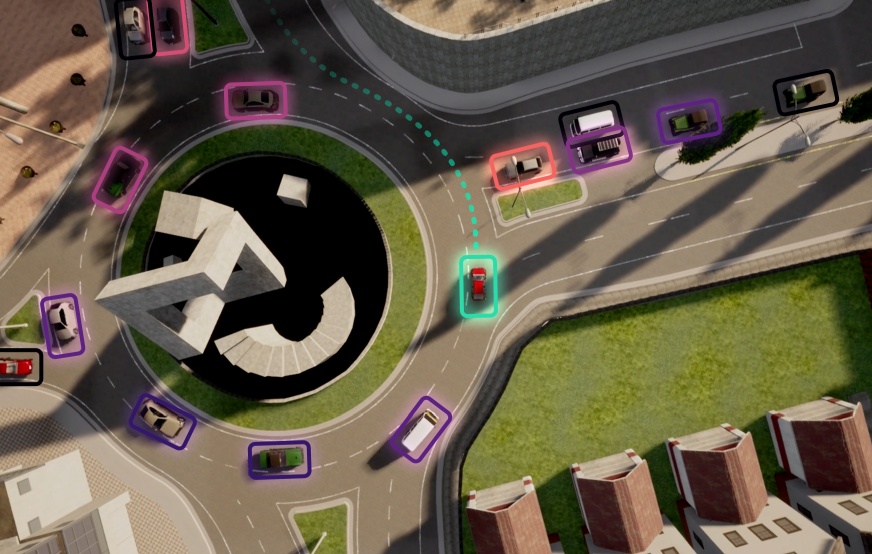}
    \caption{Traffic scene showing the roundabout scenario. The green rectangle in the center is around the ego vehicle, and the dotted green line visualizes the currently chosen action for the agent.
    All other cars are surrounded by a rectangle colored according to their importance for the risk estimate, as assigned through integrated gradients,
    high importance is colored orange, unimportant dark purple or black.}
    \label{fig:risk-interpretation}
    \vspace*{-5mm}
\end{figure}

\subsection{Risk Interpretation}
An additional benefit of the explicit disentanglement of risk and utility estimates lies in the possibility of independently explaining the risk estimator. We demonstrate this by using integrated gradients (IG)~\cite{sundararajan2017axiomatic} to assign a saliency (or "importance") for the risk estimate to individual elements of the observation vector $o$. IG estimates the saliency of an input element by summing up the gradients when stepping from a baseline (we use the zero vector) to the real input value. As IG expects a single individual estimate (instead of a vector of per-action risk estimates), we sum up the risk estimates for all actions into one combined situative risk assessment for the visualizations. Alternatively, an expert could assess the risk estimator for each action separately.

As we use an interpretable, semantic observation space, we can use IG to calculate the importance of individual semantic elements to our overall risk estimate. Here, we are most interested in how the risk estimator uses information about neighboring vehicles for its prediction. Because non-agent vehicles are represented by multiple real-valued elements in our observation vector, we can sum up the contribution of all of these elements to calculate the per-car risk saliency metric, see Fig.~\ref{fig:risk-interpretation}. Visualizations like this can be very effective in the assessment of cars that the agent considers most important for its current situational risk estimate.

When we visualized the per-car risk saliency, we found that many of our agents sensibly base their risk estimates on cars that are close to the agent and that potentially cross the agent's path.
Additionally, we became aware that both estimators were relatively sensitive to small variations in the observation. This can be very sensible (e.g., in scenarios where a small difference in position applies to a large difference in predicted trajectory), but could also indicate a potential source of problems and wrong decisions. We leave an investigation of possible remedies for future work.

\section{Conclusion}
\label{sec:conclusion}
\glsresetall%
This work introduces SafeDQN, a novel algorithm that combines explicit risk estimation and Lagrangian learning to find optimal solutions for \glspl{CMDP}. SafeDQN learns an optimal, constraint-satisfying trade-off without requiring manual tuning of hyper-parameter, in particular to the reward function. SafeDQN outperforms baselines in terms of both safety and average return across multiple traffic scenarios. In addition, SafeDQN's separated risk and utility estimators allow for independent interpretation, bootstrapping, and training.

Our findings on interpretability highlight the need for an independent assessment of safety components. We found that the risk estimators show a high-cost recall and cost precision, and use overall sensible information on nearby cars to make their decisions.
This conclusion could only be drawn by SafeDQN's unique structure and explicit safety components.

\section*{Acknowledgements}
This work was supported by the Bavarian Ministry for Economic Affairs, Infrastructure, Transport and Technology through the Center for Analytics-Data-Applications (ADA-Center) within the framework of “BAYERN DIGITAL II”. B.M.E. gratefully acknowledges support of the German Research Foundation (DFG) within the framework of the Heisenberg professorship program (Grant ES 434/8-1).
 
\bibliographystyle{IEEEtran}
\bibliography{references}

\end{document}

%% file: bin/overview.pdf_tex
\begingroup%
  \makeatletter%
  \providecommand\color[2][]{%
    \errmessage{(Inkscape) Color is used for the text in Inkscape, but the package 'color.sty' is not loaded}%
    \renewcommand\color[2][]{}%
  }%
  \providecommand\transparent[1]{%
    \errmessage{(Inkscape) Transparency is used (non-zero) for the text in Inkscape, but the package 'transparent.sty' is not loaded}%
    \renewcommand\transparent[1]{}%
  }%
  \providecommand\rotatebox[2]{#2}%
  \newcommand*\fsize{\dimexpr\f@size pt\relax}%
  \newcommand*\lineheight[1]{\fontsize{\fsize}{#1\fsize}\selectfont}%
  \ifx\svgwidth\undefined%
    \setlength{\unitlength}{901.26316821bp}%
    \ifx\svgscale\undefined%
      \relax%
    \else%
      \setlength{\unitlength}{\unitlength * \real{\svgscale}}%
    \fi%
  \else%
    \setlength{\unitlength}{\svgwidth}%
  \fi%
  \global\let\svgwidth\undefined%
  \global\let\svgscale\undefined%
  \makeatother%
  \begin{picture}(1,0.52380831)%
    \lineheight{1}%
    \setlength\tabcolsep{0pt}%
    \put(0,0){\includegraphics[width=\unitlength,page=1]{overview.pdf}}%
    \put(0.41480068,0.38807331){\color[rgb]{0,0,0}\makebox(0,0)[t]{\lineheight{10}\smash{\begin{tabular}[t]{c}Risk\end{tabular}}}}%
    \put(0,0){\includegraphics[width=\unitlength,page=2]{overview.pdf}}%
    \put(0.17249374,0.3922426){\color[rgb]{0,0,0}\makebox(0,0)[t]{\lineheight{10}\smash{\begin{tabular}[t]{c}Utility\end{tabular}}}}%
    \put(0,0){\includegraphics[width=\unitlength,page=3]{overview.pdf}}%
    \put(0.61754256,0.18227495){\color[rgb]{0,0,0}\makebox(0,0)[t]{\lineheight{10}\smash{\begin{tabular}[t]{c}Tradeoff\end{tabular}}}}%
    \put(0,0){\includegraphics[width=\unitlength,page=4]{overview.pdf}}%
    \put(0.61864239,0.1316449){\color[rgb]{0,0,0}\makebox(0,0)[t]{\lineheight{10}\smash{\begin{tabular}[t]{c}\normalsize $\lambda$\end{tabular}}}}%
    \put(0,0){\includegraphics[width=\unitlength,page=5]{overview.pdf}}%
    \put(0.27722236,0.07264339){\color[rgb]{0,0,0}\makebox(0,0)[t]{\lineheight{10}\smash{\begin{tabular}[t]{c}$\mathrm{argmax}_{a} Q(s,a)+\lambda Q_C(s,a)$\end{tabular}}}}%
    \put(0,0){\includegraphics[width=\unitlength,page=6]{overview.pdf}}%
    \put(0.5830578,0.04791623){\color[rgb]{0,0,0}\makebox(0,0)[lt]{\lineheight{10}\smash{\begin{tabular}[t]{l}Action $a$\end{tabular}}}}%
    \put(0.58176835,0.48052327){\color[rgb]{0,0,0}\makebox(0,0)[lt]{\lineheight{10}\smash{\begin{tabular}[t]{l}State $s$\end{tabular}}}}%
    \put(0.17403636,0.30677931){\color[rgb]{0,0,0}\makebox(0,0)[t]{\lineheight{10}\smash{\begin{tabular}[t]{c}\normalsize $Q$\end{tabular}}}}%
    \put(0.41634335,0.3044017){\color[rgb]{0,0,0}\makebox(0,0)[t]{\lineheight{10}\smash{\begin{tabular}[t]{c}\normalsize $Q_C$\end{tabular}}}}%
    \put(0.82831841,0.26190958){\color[rgb]{0,0,0}\rotatebox{90}{\makebox(0,0)[t]{\lineheight{10}\smash{\begin{tabular}[t]{c}{\normalsize Environment}\end{tabular}}}}}%
    \put(0.04031745,0.17951655){\color[rgb]{0,0,0}\rotatebox{90}{\makebox(0,0)[lt]{\lineheight{10}\smash{\begin{tabular}[t]{l}{\normalsize Agent}\end{tabular}}}}}%
    \put(0.17627263,0.3536703){\color[rgb]{0,0,0}\makebox(0,0)[t]{\lineheight{10}\smash{\begin{tabular}[t]{c}estimate\end{tabular}}}}%
    \put(0.41806808,0.3536703){\color[rgb]{0,0,0}\makebox(0,0)[t]{\lineheight{10}\smash{\begin{tabular}[t]{c}estimate\end{tabular}}}}%
    \put(0,0){\includegraphics[width=\unitlength,page=7]{overview.pdf}}%
  \end{picture}%
\endgroup%

%% file: bin/roadmaps.pdf_tex
\begingroup%
  \makeatletter%
  \providecommand\color[2][]{%
    \errmessage{(Inkscape) Color is used for the text in Inkscape, but the package 'color.sty' is not loaded}%
    \renewcommand\color[2][]{}%
  }%
  \providecommand\transparent[1]{%
    \errmessage{(Inkscape) Transparency is used (non-zero) for the text in Inkscape, but the package 'transparent.sty' is not loaded}%
    \renewcommand\transparent[1]{}%
  }%
  \providecommand\rotatebox[2]{#2}%
  \newcommand*\fsize{\dimexpr\f@size pt\relax}%
  \newcommand*\lineheight[1]{\fontsize{\fsize}{#1\fsize}\selectfont}%
  \ifx\svgwidth\undefined%
    \setlength{\unitlength}{725.18935316bp}%
    \ifx\svgscale\undefined%
      \relax%
    \else%
      \setlength{\unitlength}{\unitlength * \real{\svgscale}}%
    \fi%
  \else%
    \setlength{\unitlength}{\svgwidth}%
  \fi%
  \global\let\svgwidth\undefined%
  \global\let\svgscale\undefined%
  \makeatother%
  \begin{picture}(1,0.23453011)%
    \lineheight{1}%
    \setlength\tabcolsep{0pt}%
    \put(0,0){\includegraphics[width=\unitlength,page=1]{roadmaps.pdf}}%
    \put(0.03730556,0.0054863){\color[rgb]{0,0,0}\makebox(0,0)[lt]{\lineheight{1.25}\smash{\begin{tabular}[t]{l}(a) Left turn\end{tabular}}}}%
    \put(0.28324496,0.0054863){\color[rgb]{0,0,0}\makebox(0,0)[lt]{\lineheight{1.25}\smash{\begin{tabular}[t]{l}(b) Right turn\end{tabular}}}}%
    \put(0,0){\includegraphics[width=\unitlength,page=2]{roadmaps.pdf}}%
    \put(0.51596687,0.0054863){\color[rgb]{0,0,0}\makebox(0,0)[lt]{\lineheight{1.25}\smash{\begin{tabular}[t]{l}(c) Roundabout\end{tabular}}}}%
    \put(0,0){\includegraphics[width=\unitlength,page=3]{roadmaps.pdf}}%
    \put(0.89465121,0.15589361){\color[rgb]{0,0,0}\makebox(0,0)[t]{\lineheight{1.25}\smash{\begin{tabular}[t]{c}(d) Highway merge\end{tabular}}}}%
    \put(0.85841193,0.08659833){\color[rgb]{0,0,0}\makebox(0,0)[t]{\lineheight{1.25}\smash{\begin{tabular}[t]{c}(e) Highway drive\end{tabular}}}}%
    \put(0.80578885,0.0054863){\color[rgb]{0,0,0}\makebox(0,0)[t]{\lineheight{1.25}\smash{\begin{tabular}[t]{c}(f) Highway split\end{tabular}}}}%
  \end{picture}%
\endgroup%

%% file: bin/results-cropped-converted2.pdf_tex
\begingroup%
  \makeatletter%
  \providecommand\color[2][]{%
    \errmessage{(Inkscape) Color is used for the text in Inkscape, but the package 'color.sty' is not loaded}%
    \renewcommand\color[2][]{}%
  }%
  \providecommand\transparent[1]{%
    \errmessage{(Inkscape) Transparency is used (non-zero) for the text in Inkscape, but the package 'transparent.sty' is not loaded}%
    \renewcommand\transparent[1]{}%
  }%
  \providecommand\rotatebox[2]{#2}%
  \newcommand*\fsize{\dimexpr\f@size pt\relax}%
  \newcommand*\lineheight[1]{\fontsize{\fsize}{#1\fsize}\selectfont}%
  \ifx\svgwidth\undefined%
    \setlength{\unitlength}{839.31555176bp}%
    \ifx\svgscale\undefined%
      \relax%
    \else%
      \setlength{\unitlength}{\unitlength * \real{\svgscale}}%
    \fi%
  \else%
    \setlength{\unitlength}{\svgwidth}%
  \fi%
  \global\let\svgwidth\undefined%
  \global\let\svgscale\undefined%
  \makeatother%
  \begin{picture}(1,0.27927471)%
    \lineheight{1}%
    \setlength\tabcolsep{0pt}%
    \put(0,0){\includegraphics[width=\unitlength,page=1]{results-cropped-converted2.pdf}}%
    \put(0.02474649,0.15665277){\color[rgb]{0,0,0}\makebox(0,0)[lt]{\lineheight{1.25}\smash{\begin{tabular}[t]{l}0\end{tabular}}}}%
    \put(0,0){\includegraphics[width=\unitlength,page=2]{results-cropped-converted2.pdf}}%
    \put(0.02474649,0.19557106){\color[rgb]{0,0,0}\makebox(0,0)[lt]{\lineheight{1.25}\smash{\begin{tabular}[t]{l}5\end{tabular}}}}%
    \put(0,0){\includegraphics[width=\unitlength,page=3]{results-cropped-converted2.pdf}}%
    \put(0.01716964,0.23448935){\color[rgb]{0,0,0}\makebox(0,0)[lt]{\lineheight{1.25}\smash{\begin{tabular}[t]{l}10\end{tabular}}}}%
    \put(0,0){\includegraphics[width=\unitlength,page=4]{results-cropped-converted2.pdf}}%
    \put(0.07966362,0.26825659){\color[rgb]{0,0,0}\makebox(0,0)[lt]{\lineheight{1.25}\smash{\begin{tabular}[t]{l}Left-Turn\end{tabular}}}}%
    \put(0,0){\includegraphics[width=\unitlength,page=5]{results-cropped-converted2.pdf}}%
    \put(0.19006879,0.15665277){\color[rgb]{0,0,0}\makebox(0,0)[lt]{\lineheight{1.25}\smash{\begin{tabular}[t]{l}0\end{tabular}}}}%
    \put(0,0){\includegraphics[width=\unitlength,page=6]{results-cropped-converted2.pdf}}%
    \put(0.18249193,0.19427714){\color[rgb]{0,0,0}\makebox(0,0)[lt]{\lineheight{1.25}\smash{\begin{tabular}[t]{l}10\end{tabular}}}}%
    \put(0,0){\includegraphics[width=\unitlength,page=7]{results-cropped-converted2.pdf}}%
    \put(0.18249193,0.23190151){\color[rgb]{0,0,0}\makebox(0,0)[lt]{\lineheight{1.25}\smash{\begin{tabular}[t]{l}20\end{tabular}}}}%
    \put(0,0){\includegraphics[width=\unitlength,page=8]{results-cropped-converted2.pdf}}%
    \put(0.26941131,0.26825659){\color[rgb]{0,0,0}\makebox(0,0)[t]{\lineheight{1.25}\smash{\begin{tabular}[t]{c}Right-Turn\end{tabular}}}}%
    \put(0,0){\includegraphics[width=\unitlength,page=9]{results-cropped-converted2.pdf}}%
    \put(0.35539109,0.15613777){\color[rgb]{0,0,0}\makebox(0,0)[lt]{\lineheight{1.25}\smash{\begin{tabular}[t]{l}0\end{tabular}}}}%
    \put(0,0){\includegraphics[width=\unitlength,page=10]{results-cropped-converted2.pdf}}%
    \put(0.34781422,0.20687627){\color[rgb]{0,0,0}\makebox(0,0)[lt]{\lineheight{1.25}\smash{\begin{tabular}[t]{l}10\end{tabular}}}}%
    \put(0,0){\includegraphics[width=\unitlength,page=11]{results-cropped-converted2.pdf}}%
    \put(0.43519244,0.26825659){\color[rgb]{0,0,0}\makebox(0,0)[t]{\lineheight{1.25}\smash{\begin{tabular}[t]{c}Roundabout\end{tabular}}}}%
    \put(0,0){\includegraphics[width=\unitlength,page=12]{results-cropped-converted2.pdf}}%
    \put(0.50933878,0.15665277){\color[rgb]{0,0,0}\makebox(0,0)[lt]{\lineheight{1.25}\smash{\begin{tabular}[t]{l}0.0\end{tabular}}}}%
    \put(0,0){\includegraphics[width=\unitlength,page=13]{results-cropped-converted2.pdf}}%
    \put(0.50933878,0.18804219){\color[rgb]{0,0,0}\makebox(0,0)[lt]{\lineheight{1.25}\smash{\begin{tabular}[t]{l}2.5\end{tabular}}}}%
    \put(0,0){\includegraphics[width=\unitlength,page=14]{results-cropped-converted2.pdf}}%
    \put(0.50933878,0.21943162){\color[rgb]{0,0,0}\makebox(0,0)[lt]{\lineheight{1.25}\smash{\begin{tabular}[t]{l}5.0\end{tabular}}}}%
    \put(0,0){\includegraphics[width=\unitlength,page=15]{results-cropped-converted2.pdf}}%
    \put(0.50933878,0.25082104){\color[rgb]{0,0,0}\makebox(0,0)[lt]{\lineheight{1.25}\smash{\begin{tabular}[t]{l}7.5\end{tabular}}}}%
    \put(0,0){\includegraphics[width=\unitlength,page=16]{results-cropped-converted2.pdf}}%
    \put(0.60025394,0.26825659){\color[rgb]{0,0,0}\makebox(0,0)[t]{\lineheight{1.25}\smash{\begin{tabular}[t]{c}Highway-Merge\end{tabular}}}}%
    \put(0,0){\includegraphics[width=\unitlength,page=17]{results-cropped-converted2.pdf}}%
    \put(0.68603567,0.15575716){\color[rgb]{0,0,0}\makebox(0,0)[lt]{\lineheight{1.25}\smash{\begin{tabular}[t]{l}0\end{tabular}}}}%
    \put(0,0){\includegraphics[width=\unitlength,page=18]{results-cropped-converted2.pdf}}%
    \put(0.67845881,0.19249985){\color[rgb]{0,0,0}\makebox(0,0)[lt]{\lineheight{1.25}\smash{\begin{tabular}[t]{l}10\end{tabular}}}}%
    \put(0,0){\includegraphics[width=\unitlength,page=19]{results-cropped-converted2.pdf}}%
    \put(0.67845881,0.22924253){\color[rgb]{0,0,0}\makebox(0,0)[lt]{\lineheight{1.25}\smash{\begin{tabular}[t]{l}20\end{tabular}}}}%
    \put(0,0){\includegraphics[width=\unitlength,page=20]{results-cropped-converted2.pdf}}%
    \put(0.76559746,0.26825659){\color[rgb]{0,0,0}\makebox(0,0)[t]{\lineheight{1.25}\smash{\begin{tabular}[t]{c}Highway-Drive\end{tabular}}}}%
    \put(0,0){\includegraphics[width=\unitlength,page=21]{results-cropped-converted2.pdf}}%
    \put(0.85135796,0.15642976){\color[rgb]{0,0,0}\makebox(0,0)[lt]{\lineheight{1.25}\smash{\begin{tabular}[t]{l}0\end{tabular}}}}%
    \put(0,0){\includegraphics[width=\unitlength,page=22]{results-cropped-converted2.pdf}}%
    \put(0.85135796,0.20711286){\color[rgb]{0,0,0}\makebox(0,0)[lt]{\lineheight{1.25}\smash{\begin{tabular}[t]{l}5\end{tabular}}}}%
    \put(0,0){\includegraphics[width=\unitlength,page=23]{results-cropped-converted2.pdf}}%
    \put(0.01214846,0.20036095){\color[rgb]{0,0,0}\rotatebox{90}{\makebox(0,0)[t]{\lineheight{1.25}\smash{\begin{tabular}[t]{c}Crashes\end{tabular}}}}}%
    \put(0.93110534,0.26825659){\color[rgb]{0,0,0}\makebox(0,0)[t]{\lineheight{1.25}\smash{\begin{tabular}[t]{c}Highway-Split\end{tabular}}}}%
    \put(0,0){\includegraphics[width=\unitlength,page=24]{results-cropped-converted2.pdf}}%
    \put(0.03687505,0.0247561){\color[rgb]{0,0,0}\makebox(0,0)[lt]{\lineheight{1.25}\smash{\begin{tabular}[t]{l}0\end{tabular}}}}%
    \put(0,0){\includegraphics[width=\unitlength,page=25]{results-cropped-converted2.pdf}}%
    \put(0.10145407,0.0247561){\color[rgb]{0,0,0}\makebox(0,0)[lt]{\lineheight{1.25}\smash{\begin{tabular}[t]{l}1\end{tabular}}}}%
    \put(0,0){\includegraphics[width=\unitlength,page=26]{results-cropped-converted2.pdf}}%
    \put(0.16603309,0.0247561){\color[rgb]{0,0,0}\makebox(0,0)[lt]{\lineheight{1.25}\smash{\begin{tabular}[t]{l}2\end{tabular}}}}%
    \put(0,0){\includegraphics[width=\unitlength,page=27]{results-cropped-converted2.pdf}}%
    \put(0.02295932,0.07134457){\color[rgb]{0,0,0}\makebox(0,0)[lt]{\lineheight{1.25}\smash{\begin{tabular}[t]{l}0\end{tabular}}}}%
    \put(0,0){\includegraphics[width=\unitlength,page=28]{results-cropped-converted2.pdf}}%
    \put(0.01137995,0.11757473){\color[rgb]{0,0,0}\makebox(0,0)[lt]{\lineheight{1.25}\smash{\begin{tabular}[t]{l}100\end{tabular}}}}%
    \put(0,0){\includegraphics[width=\unitlength,page=29]{results-cropped-converted2.pdf}}%
    \put(0.20219735,0.0247561){\color[rgb]{0,0,0}\makebox(0,0)[lt]{\lineheight{1.25}\smash{\begin{tabular}[t]{l}0\end{tabular}}}}%
    \put(0,0){\includegraphics[width=\unitlength,page=30]{results-cropped-converted2.pdf}}%
    \put(0.26677637,0.0247561){\color[rgb]{0,0,0}\makebox(0,0)[lt]{\lineheight{1.25}\smash{\begin{tabular}[t]{l}1\end{tabular}}}}%
    \put(0,0){\includegraphics[width=\unitlength,page=31]{results-cropped-converted2.pdf}}%
    \put(0.33135538,0.0247561){\color[rgb]{0,0,0}\makebox(0,0)[lt]{\lineheight{1.25}\smash{\begin{tabular}[t]{l}2\end{tabular}}}}%
    \put(0,0){\includegraphics[width=\unitlength,page=32]{results-cropped-converted2.pdf}}%
    \put(0.19006879,0.03873778){\color[rgb]{0,0,0}\makebox(0,0)[lt]{\lineheight{1.25}\smash{\begin{tabular}[t]{l}0\end{tabular}}}}%
    \put(0,0){\includegraphics[width=\unitlength,page=33]{results-cropped-converted2.pdf}}%
    \put(0.17491507,0.09076134){\color[rgb]{0,0,0}\makebox(0,0)[lt]{\lineheight{1.25}\smash{\begin{tabular}[t]{l}100\end{tabular}}}}%
    \put(0,0){\includegraphics[width=\unitlength,page=34]{results-cropped-converted2.pdf}}%
    \put(0.36751963,0.0247561){\color[rgb]{0,0,0}\makebox(0,0)[lt]{\lineheight{1.25}\smash{\begin{tabular}[t]{l}0\end{tabular}}}}%
    \put(0,0){\includegraphics[width=\unitlength,page=35]{results-cropped-converted2.pdf}}%
    \put(0.43209866,0.0247561){\color[rgb]{0,0,0}\makebox(0,0)[lt]{\lineheight{1.25}\smash{\begin{tabular}[t]{l}1\end{tabular}}}}%
    \put(0,0){\includegraphics[width=\unitlength,page=36]{results-cropped-converted2.pdf}}%
    \put(0.49667768,0.0247561){\color[rgb]{0,0,0}\makebox(0,0)[lt]{\lineheight{1.25}\smash{\begin{tabular}[t]{l}2\end{tabular}}}}%
    \put(0,0){\includegraphics[width=\unitlength,page=37]{results-cropped-converted2.pdf}}%
    \put(0.34781422,0.07564548){\color[rgb]{0,0,0}\makebox(0,0)[lt]{\lineheight{1.25}\smash{\begin{tabular}[t]{l}50\end{tabular}}}}%
    \put(0,0){\includegraphics[width=\unitlength,page=38]{results-cropped-converted2.pdf}}%
    \put(0.34023737,0.11466401){\color[rgb]{0,0,0}\makebox(0,0)[lt]{\lineheight{1.25}\smash{\begin{tabular}[t]{l}100\end{tabular}}}}%
    \put(0,0){\includegraphics[width=\unitlength,page=39]{results-cropped-converted2.pdf}}%
    \put(0.53284193,0.0247561){\color[rgb]{0,0,0}\makebox(0,0)[lt]{\lineheight{1.25}\smash{\begin{tabular}[t]{l}0\end{tabular}}}}%
    \put(0,0){\includegraphics[width=\unitlength,page=40]{results-cropped-converted2.pdf}}%
    \put(0.59742096,0.0247561){\color[rgb]{0,0,0}\makebox(0,0)[lt]{\lineheight{1.25}\smash{\begin{tabular}[t]{l}1\end{tabular}}}}%
    \put(0,0){\includegraphics[width=\unitlength,page=41]{results-cropped-converted2.pdf}}%
    \put(0.66199998,0.0247561){\color[rgb]{0,0,0}\makebox(0,0)[lt]{\lineheight{1.25}\smash{\begin{tabular}[t]{l}2\end{tabular}}}}%
    \put(0,0){\includegraphics[width=\unitlength,page=42]{results-cropped-converted2.pdf}}%
    \put(0.51314117,0.05507669){\color[rgb]{0,0,0}\makebox(0,0)[lt]{\lineheight{1.25}\smash{\begin{tabular}[t]{l}50\end{tabular}}}}%
    \put(0,0){\includegraphics[width=\unitlength,page=43]{results-cropped-converted2.pdf}}%
    \put(0.52071337,0.09046701){\color[rgb]{0,0,0}\makebox(0,0)[lt]{\lineheight{1.25}\smash{\begin{tabular}[t]{l}0\end{tabular}}}}%
    \put(0,0){\includegraphics[width=\unitlength,page=44]{results-cropped-converted2.pdf}}%
    \put(0.51313652,0.12585732){\color[rgb]{0,0,0}\makebox(0,0)[lt]{\lineheight{1.25}\smash{\begin{tabular}[t]{l}50\end{tabular}}}}%
    \put(0,0){\includegraphics[width=\unitlength,page=45]{results-cropped-converted2.pdf}}%
    \put(0.69816422,0.0247561){\color[rgb]{0,0,0}\makebox(0,0)[lt]{\lineheight{1.25}\smash{\begin{tabular}[t]{l}0\end{tabular}}}}%
    \put(0,0){\includegraphics[width=\unitlength,page=46]{results-cropped-converted2.pdf}}%
    \put(0.76274324,0.0247561){\color[rgb]{0,0,0}\makebox(0,0)[lt]{\lineheight{1.25}\smash{\begin{tabular}[t]{l}1\end{tabular}}}}%
    \put(0,0){\includegraphics[width=\unitlength,page=47]{results-cropped-converted2.pdf}}%
    \put(0.82732227,0.0247561){\color[rgb]{0,0,0}\makebox(0,0)[lt]{\lineheight{1.25}\smash{\begin{tabular}[t]{l}2\end{tabular}}}}%
    \put(0,0){\includegraphics[width=\unitlength,page=48]{results-cropped-converted2.pdf}}%
    \put(0.68603567,0.05571029){\color[rgb]{0,0,0}\makebox(0,0)[lt]{\lineheight{1.25}\smash{\begin{tabular}[t]{l}0\end{tabular}}}}%
    \put(0,0){\includegraphics[width=\unitlength,page=49]{results-cropped-converted2.pdf}}%
    \put(0.67845881,0.08451043){\color[rgb]{0,0,0}\makebox(0,0)[lt]{\lineheight{1.25}\smash{\begin{tabular}[t]{l}50\end{tabular}}}}%
    \put(0,0){\includegraphics[width=\unitlength,page=50]{results-cropped-converted2.pdf}}%
    \put(0.67088195,0.11331058){\color[rgb]{0,0,0}\makebox(0,0)[lt]{\lineheight{1.25}\smash{\begin{tabular}[t]{l}100\end{tabular}}}}%
    \put(0,0){\includegraphics[width=\unitlength,page=51]{results-cropped-converted2.pdf}}%
    \put(0.67088195,0.14211072){\color[rgb]{0,0,0}\makebox(0,0)[lt]{\lineheight{1.25}\smash{\begin{tabular}[t]{l}150\end{tabular}}}}%
    \put(0,0){\includegraphics[width=\unitlength,page=52]{results-cropped-converted2.pdf}}%
    \put(0.86348652,0.0247561){\color[rgb]{0,0,0}\makebox(0,0)[lt]{\lineheight{1.25}\smash{\begin{tabular}[t]{l}0\end{tabular}}}}%
    \put(0,0){\includegraphics[width=\unitlength,page=53]{results-cropped-converted2.pdf}}%
    \put(0.92806554,0.0247561){\color[rgb]{0,0,0}\makebox(0,0)[lt]{\lineheight{1.25}\smash{\begin{tabular}[t]{l}1\end{tabular}}}}%
    \put(0,0){\includegraphics[width=\unitlength,page=54]{results-cropped-converted2.pdf}}%
    \put(0.99264457,0.0247561){\color[rgb]{0,0,0}\makebox(0,0)[lt]{\lineheight{1.25}\smash{\begin{tabular}[t]{l}2\end{tabular}}}}%
    \put(0,0){\includegraphics[width=\unitlength,page=55]{results-cropped-converted2.pdf}}%
    \put(0.85135796,0.06900574){\color[rgb]{0,0,0}\makebox(0,0)[lt]{\lineheight{1.25}\smash{\begin{tabular}[t]{l}0\end{tabular}}}}%
    \put(0,0){\includegraphics[width=\unitlength,page=56]{results-cropped-converted2.pdf}}%
    \put(0.83620424,0.11242729){\color[rgb]{0,0,0}\makebox(0,0)[lt]{\lineheight{1.25}\smash{\begin{tabular}[t]{l}100\end{tabular}}}}%
    \put(0,0){\includegraphics[width=\unitlength,page=57]{results-cropped-converted2.pdf}}%
    \put(0.01175866,0.10575623){\color[rgb]{0,0,0}\rotatebox{90}{\makebox(0,0)[rt]{\lineheight{1.25}\smash{\begin{tabular}[t]{r}Return\end{tabular}}}}}%
    \put(0,0){\includegraphics[width=\unitlength,page=58]{results-cropped-converted2.pdf}}%
    \put(0.27218024,0.00265699){\color[rgb]{0,0,0}\makebox(0,0)[lt]{\lineheight{1.25}\smash{\begin{tabular}[t]{l}DQN\end{tabular}}}}%
    \put(0,0){\includegraphics[width=\unitlength,page=59]{results-cropped-converted2.pdf}}%
    \put(0.35684744,0.00265699){\color[rgb]{0,0,0}\makebox(0,0)[lt]{\lineheight{1.25}\smash{\begin{tabular}[t]{l}PPO\end{tabular}}}}%
    \put(0,0){\includegraphics[width=\unitlength,page=60]{results-cropped-converted2.pdf}}%
    \put(0.43779137,0.00265699){\color[rgb]{0,0,0}\makebox(0,0)[lt]{\lineheight{1.25}\smash{\begin{tabular}[t]{l}RCPO+\end{tabular}}}}%
    \put(0,0){\includegraphics[width=\unitlength,page=61]{results-cropped-converted2.pdf}}%
    \put(0.53753782,0.00265699){\color[rgb]{0,0,0}\makebox(0,0)[lt]{\lineheight{1.25}\smash{\begin{tabular}[t]{l}SafeDQN\end{tabular}}}}%
    \put(0,0){\includegraphics[width=\unitlength,page=62]{results-cropped-converted2.pdf}}%
    \put(0.64858441,0.00265699){\color[rgb]{0,0,0}\makebox(0,0)[lt]{\lineheight{1.25}\smash{\begin{tabular}[t]{l}SafeDQN-AltObjective\end{tabular}}}}%
    \put(-0.15894914,-0.03091298){\color[rgb]{0,0,0}\makebox(0,0)[lt]{\begin{minipage}{0.18925303\unitlength}\centering \end{minipage}}}%
    \put(0.05328637,0.00345663){\color[rgb]{0,0,0}\makebox(0,0)[lt]{\lineheight{1.25}\smash{\begin{tabular}[t]{l}Time steps (1e6)\end{tabular}}}}%
  \end{picture}%
\endgroup%

%% file: bin/recall-precision-converted2.pdf_tex
\begingroup%
  \makeatletter%
  \providecommand\color[2][]{%
    \errmessage{(Inkscape) Color is used for the text in Inkscape, but the package 'color.sty' is not loaded}%
    \renewcommand\color[2][]{}%
  }%
  \providecommand\transparent[1]{%
    \errmessage{(Inkscape) Transparency is used (non-zero) for the text in Inkscape, but the package 'transparent.sty' is not loaded}%
    \renewcommand\transparent[1]{}%
  }%
  \providecommand\rotatebox[2]{#2}%
  \newcommand*\fsize{\dimexpr\f@size pt\relax}%
  \newcommand*\lineheight[1]{\fontsize{\fsize}{#1\fsize}\selectfont}%
  \ifx\svgwidth\undefined%
    \setlength{\unitlength}{302.1620636bp}%
    \ifx\svgscale\undefined%
      \relax%
    \else%
      \setlength{\unitlength}{\unitlength * \real{\svgscale}}%
    \fi%
  \else%
    \setlength{\unitlength}{\svgwidth}%
  \fi%
  \global\let\svgwidth\undefined%
  \global\let\svgscale\undefined%
  \makeatother%
  \begin{picture}(1,0.75863531)%
    \lineheight{1}%
    \setlength\tabcolsep{0pt}%
    \put(0,0){\includegraphics[width=\unitlength,page=1]{recall-precision-converted2.pdf}}%
    \put(-0.00218154,0.44943189){\color[rgb]{0,0,0}\makebox(0,0)[lt]{\lineheight{1.25}\smash{\begin{tabular}[t]{l}0.0\end{tabular}}}}%
    \put(0,0){\includegraphics[width=\unitlength,page=2]{recall-precision-converted2.pdf}}%
    \put(-0.00218154,0.5917523){\color[rgb]{0,0,0}\makebox(0,0)[lt]{\lineheight{1.25}\smash{\begin{tabular}[t]{l}0.5\end{tabular}}}}%
    \put(0,0){\includegraphics[width=\unitlength,page=3]{recall-precision-converted2.pdf}}%
    \put(-0.00218154,0.73407274){\color[rgb]{0,0,0}\makebox(0,0)[lt]{\lineheight{1.25}\smash{\begin{tabular}[t]{l}1.0\end{tabular}}}}%
    \put(0,0){\includegraphics[width=\unitlength,page=4]{recall-precision-converted2.pdf}}%
    \put(0.46915904,0.68903457){\color[rgb]{0,0,0}\makebox(0,0)[lt]{\lineheight{1.25}\smash{\begin{tabular}[t]{l}Crash rate per episode\end{tabular}}}}%
    \put(0,0){\includegraphics[width=\unitlength,page=5]{recall-precision-converted2.pdf}}%
    \put(0.0769971,0.05213076){\color[rgb]{0,0,0}\makebox(0,0)[lt]{\lineheight{1.25}\smash{\begin{tabular}[t]{l}0.00\end{tabular}}}}%
    \put(0,0){\includegraphics[width=\unitlength,page=6]{recall-precision-converted2.pdf}}%
    \put(0.18234554,0.05213076){\color[rgb]{0,0,0}\makebox(0,0)[lt]{\lineheight{1.25}\smash{\begin{tabular}[t]{l}0.25\end{tabular}}}}%
    \put(0,0){\includegraphics[width=\unitlength,page=7]{recall-precision-converted2.pdf}}%
    \put(0.28769398,0.05213076){\color[rgb]{0,0,0}\makebox(0,0)[lt]{\lineheight{1.25}\smash{\begin{tabular}[t]{l}0.50\end{tabular}}}}%
    \put(0,0){\includegraphics[width=\unitlength,page=8]{recall-precision-converted2.pdf}}%
    \put(0.39304241,0.05213076){\color[rgb]{0,0,0}\makebox(0,0)[lt]{\lineheight{1.25}\smash{\begin{tabular}[t]{l}0.75\end{tabular}}}}%
    \put(0,0){\includegraphics[width=\unitlength,page=9]{recall-precision-converted2.pdf}}%
    \put(0.49839088,0.05213076){\color[rgb]{0,0,0}\makebox(0,0)[lt]{\lineheight{1.25}\smash{\begin{tabular}[t]{l}1.00\end{tabular}}}}%
    \put(0,0){\includegraphics[width=\unitlength,page=10]{recall-precision-converted2.pdf}}%
    \put(0.60373931,0.05213076){\color[rgb]{0,0,0}\makebox(0,0)[lt]{\lineheight{1.25}\smash{\begin{tabular}[t]{l}1.25\end{tabular}}}}%
    \put(0,0){\includegraphics[width=\unitlength,page=11]{recall-precision-converted2.pdf}}%
    \put(0.70908774,0.05213076){\color[rgb]{0,0,0}\makebox(0,0)[lt]{\lineheight{1.25}\smash{\begin{tabular}[t]{l}1.50\end{tabular}}}}%
    \put(0,0){\includegraphics[width=\unitlength,page=12]{recall-precision-converted2.pdf}}%
    \put(0.8144362,0.05213076){\color[rgb]{0,0,0}\makebox(0,0)[lt]{\lineheight{1.25}\smash{\begin{tabular}[t]{l}1.75\end{tabular}}}}%
    \put(0,0){\includegraphics[width=\unitlength,page=13]{recall-precision-converted2.pdf}}%
    \put(0.91978463,0.05213076){\color[rgb]{0,0,0}\makebox(0,0)[lt]{\lineheight{1.25}\smash{\begin{tabular}[t]{l}2.00\end{tabular}}}}%
    \put(0.44395112,0.00688394){\color[rgb]{0,0,0}\makebox(0,0)[lt]{\lineheight{1.25}\smash{\begin{tabular}[t]{l}Time  steps\end{tabular}}}}%
    \put(0.93620973,0.01019342){\color[rgb]{0,0,0}\makebox(0,0)[lt]{\lineheight{1.25}\smash{\begin{tabular}[t]{l}1e6\end{tabular}}}}%
    \put(0,0){\includegraphics[width=\unitlength,page=14]{recall-precision-converted2.pdf}}%
    \put(-0.00218154,0.09100947){\color[rgb]{0,0,0}\makebox(0,0)[lt]{\lineheight{1.25}\smash{\begin{tabular}[t]{l}0.0\end{tabular}}}}%
    \put(0,0){\includegraphics[width=\unitlength,page=15]{recall-precision-converted2.pdf}}%
    \put(-0.00218154,0.23289513){\color[rgb]{0,0,0}\makebox(0,0)[lt]{\lineheight{1.25}\smash{\begin{tabular}[t]{l}0.5\end{tabular}}}}%
    \put(0,0){\includegraphics[width=\unitlength,page=16]{recall-precision-converted2.pdf}}%
    \put(-0.00218154,0.37478079){\color[rgb]{0,0,0}\makebox(0,0)[lt]{\lineheight{1.25}\smash{\begin{tabular}[t]{l}1.0\end{tabular}}}}%
    \put(0,0){\includegraphics[width=\unitlength,page=17]{recall-precision-converted2.pdf}}%
    \put(0.46879706,0.18068346){\color[rgb]{0,0,0}\makebox(0,0)[lt]{\lineheight{1.25}\smash{\begin{tabular}[t]{l}Cost Recall\end{tabular}}}}%
    \put(0,0){\includegraphics[width=\unitlength,page=18]{recall-precision-converted2.pdf}}%
    \put(0.46879706,0.13212715){\color[rgb]{0,0,0}\makebox(0,0)[lt]{\lineheight{1.25}\smash{\begin{tabular}[t]{l}Cost Precision\end{tabular}}}}%
  \end{picture}%
\endgroup%